\documentclass[journal]{IEEEtran}
\usepackage{graphicx}
\usepackage{amsmath, amssymb}
\usepackage{caption, subcaption}
\usepackage{float}
\usepackage{multirow}
\usepackage{hyperref}
\usepackage{soul,color}
\usepackage{cite}
\begin{document}
\title{Sharpend Cosine Similarity based Neural Network for Hyperspectral Image Classification}
\author{Muhammad Ahmad 
\thanks{M. Ahmad is with the Department of Computer Science, National University of Computer and Emerging Sciences, Chiniot 35400, Pakistan. E-mail: mahmad00@gmail.com}
}
\markboth{May~2023}
{M. Ahmad \MakeLowercase{\textit{et al.}}: SCS for HSIC}
\maketitle
\begin{abstract}
Hyperspectral Image Classification (HSIC) is a difficult task due to high inter and intra-class similarity and variability, nested regions, and overlapping. 2D Convolutional Neural Networks (CNN) emerged as a viable network whereas, 3D CNNs are a better alternative due to accurate classification. However, 3D CNNs are highly computationally complex due to their volume and spectral dimensions. Moreover, down-sampling and hierarchical filtering (high frequency) i.e., texture features need to be smoothed during the forward pass which is crucial for accurate HSIC. Furthermore, CNN requires tons of tuning parameters which increases the training time. Therefore, to overcome the aforesaid issues, Sharpened Cosine Similarity (SCS) concept as an alternative to convolutions in a Neural Network for HSIC is introduced. SCS is exceptionally parameter efficient due to skipping the non-linear activation layers, normalization, and dropout after the SCS layer. Use of MaxAbsPool instead of MaxPool which selects the element with the highest magnitude of activity, even if it's negative. Experimental results on publicly available HSI datasets proved the performance of SCS as compared to the convolutions in Neural Networks.
\end{abstract}
\IEEEpeerreviewmaketitle
\section{Introduction}
\label{Sec.1}

Hyperspectral Image Classification (HSIC) has been an extensively studied area of research for decades for many applications \cite{9205804, khan2021hyperspectral, app10217783, 9870277, s21093045}. HSIC task requires assigning a unique class label to a set of pixels according to the information presented in HSI \cite{9832932, Ahmad2022HyperspectralIC, roy2022hyperspectral}. Early HSIC approaches are based on handcrafted features which are extracted from spectral (color, intensity, etc.) or spatial (texture, shape, etc.) information or their combination. Among the handcrafted features, HoG, texture, and color features are widely employed. However, the classification performance largely deteriorated since these approaches are not able to extract rich semantic features from HSI.

To overcome the aforesaid limitations, several unsupervised feature-learning techniques have been proposed. Such techniques focus on learning a set of basis functions used for feature encoding for an image \cite{9832932}. These approaches can learn more discriminative information thus suitable for representing the information. However, the discriminative performance is still limited as these techniques do not make use of class information, hence, do not well represent distinguishable features among different classes \cite{9903062}. 

To overcome the aforesaid limitations, deep learning approaches have been proposed, especially Convolutional Neural Network (CNN) \cite{YU201788, hong2021graph, 8390931, ahmad2020fast}. CNNs are able to use class information to learn discriminative features, thus the features learned by CNN are more robust and discriminative for HSIC. However, due to the down-sampling and hierarchical filtering processes, the high frequency (i.e., texture features) gradually needs to be smoothened during the forward processes \cite{9411883}. Thus, to some extent, CNN architecture can be considered a low-frequency pathway for HSIC. Moreover, traditional CNN models require tons of parameters to train the model which increases the training time. Therefore, to overcome the aforesaid issues, this work proposed the use of the Sharpened Cosine Similarity (SCS) concept as an alternative to convolutions in a neural network. More specifically, the following benefits can be obtained using SCS instead of convolutions in neural networks. SCS appears to be parameter efficiency and architecture simplicity due to the nonexistence of nonlinear activation layers, dropout, and normalization layers, like batch normalization or layer normalization, after SCS layers. Use AbsMaxpool instead of Maxpool. It selects the element with the highest magnitude of activity, even if it's negative.

\section{Literature Review}
\label{LW}

High-level earth observation using HSI has become cutting-edge research. Hence HSIC has drawn widespread attention and brought several state-of-the-art methodologies since its emergence \cite{9875399}. However, most of the early approaches only considered shallow processing which requires feature engineering prior to the classification. Moreover, shallow-level processing requires extensive parameter tuning, settings, and experience which affects the learning capabilities. Therefore, during the past decade, deep learning such as CNNs and similar methodologies have been proposed. 

The works \cite{8445697, 8509610} proposed an HSIC method based on deep pyramid residual and capsule networks. Similarly, the work \cite{8061020} proposed a spectral-spatial residual network in which the residual blocks were used for identity mapping to connect 3D CLs whereas the work \cite{8082108} proposed an unsupervised HSIC method to better explore the residual CLs. The work \cite{7514991} proposed a 3D CNN network for both spatial-spectral feature learning and classification. The work \cite{9170817} proposed a mini-batch Graph CNN (miniGCN) which addresses the complexity of graph computation and feature fusion extracted by CNN and miniGCN as an end-to-end network. 

From the above discussion, one can conclude that features learned by CNN are more robust and discriminative for HSIC as compared to the other methods. However, due to the down-sampling and hierarchical filtering process, the high-frequency features gradually need to be smoothened during the forward pass. To some extent, these features are considered crucial for accurate classification, therefore, CNN architecture is referred to as a low-frequency pathway. Moreover, parameter explosion is yet another issue that increases the computational time. To overcome the aforesaid issues, this work implemented an SCS as an alternative to Convolutional Layers (CLs) for HSIC. There are a number of benefits that can be obtained using SCS instead of CLs in neural networks. 

For instance, SCS appears to be parameter efficient, and its architecture simplicity. It's not about the accuracy records but it's killing in parameter efficiency leaderboard. Skip the non-linear activation layers, dropout layers, and normalization layers after SCS. SCS uses Abs Max-pool instead of Max-pool which selects the element with the highest magnitude of activity even if it's negative. Thus, in a nutshell, the SCS works better in terms of the computational cost requires to train a DNN as compared to deep CNNs.

\section{Methodology}
\label{Meth}

Let us consider an HSI $\textbf{X} \in \mathcal{R}^{(M\times{N})\times{B}}$ where $(M\times N)$ represents the spatial region over the Earth's surface and $B$ refers to be the number of spectral bands exists in HSI. All $x_{ij} \in \textbf{X}$ pixels of HSI are classified into $C$ disjoint land-cover classes denoted as $Y = (y_1, y_2, \dots, y_n)$. Each $x_{ij} \in \textbf{X}$ (where $i = 1, 2, \dots, M$ and $j = 1, 2, \dots, N$) define a land-cover pixel as a spectral vector $x_{ij} = [x_{i,j,1}, x_{i,j,2}, \dots, x_{i,j,B}] \in \textbf{X}$ contain $B$ number of values.

Moreover, to process the spatial information, the patch extraction process is carried out as a preprocessing step where the HSI cube $\textbf{x}_{i,j} \in \mathcal{R}^{(k \times k) \times B}$ with the neighboring regions of size $k \times k$ centered at targeted pixel $(i,j)$. joint spectral-spatial features can increase the discriminative power of any model, thus the spectral-spatial cubes $\textbf{x}_{i,j} \in \mathcal{R}^{(k \times k) \times B}$ are extracted from raw data and stacked into $X$ before feature extraction. Finally, the training and test samples were selected for each class.

Convolution in Neural Network is a sliding dot product between the patch of the image and the kernel aligned which may not be a good similarity metric and thus may skip some important information which makes it an inappropriate feature detector. However, normalizing both to a magnitude of $1$, thus it will turn to cosine similarity. The cosine of two non-zero vectors can be derived by using the Euclidean distance as $k \cdot x_i = ||k||~||x_i|| cos\theta$, given $k$ and $x_i$, the cosine similarity $cos(\theta)$ can be computed as $cos(\theta) = \frac{k \cdot x_i}{||k||~||x_i||} = \frac{\sum_{i=1}^{n} k_i~ x_i}{\sqrt{\sum_{i=1}^{n} k_i^2}\sqrt{\sum_{i=1}^{n} x_i^2}}$. 

The resulting similarity ranges from $-1$ to $1$, where $-1$ means that the kernel and the image patch are exactly opposite and $0.5$ means intermediate similarity or dissimilarity among the kernel and image, whereas, $1$ means that the kernel and the patch are aligned. The main issue in cosine similarity is a very small magnitude between the kernel and the image patch which in principle desired to be magnitude in variance. However, keeping it to an extreme may end up gathering background or noise rather than foreground information. Thus, extra parameters can help to overcome the aforesaid issues. 

Therefore, the SCS was initiated, which is a stridden operation similar to convolution that extracts features from an image patch. SCS is similar to the convolutional process, however, have some important differences. In practice, the convolution is a stridden dot product among the kernel $k$ and the image patch $x_i$ as $k \cdot x_i$, whereas in cosine similarity, the image patch and kernel are normalized to have a magnitude of $1$ before the dot product is taken. It is so named because, in two dimensions, it gives the cosine of the angle between the image patch and the kernel as $\frac{k \cdot x_i}{||k||~||x_j||}$. Therefore, the cosine can be sharpened by raising the magnitude of the result to a power $p$ while maintaining the sign as $sign(k \cdot x_i) \left\vert \frac{k \cdot x_i}{||k||~||x_i||} \right\vert^{p}$. This measure can become numerically unstable if ever the magnitude of the image patch or kernel gets too close to zero. In practice, the kernel magnitude doesn't get too small and doesn't need $q$, however, maybe a good practice to have a small constraint on it as well, i.e., $sign(k \cdot x_i) \left\vert \frac{k \cdot x_i}{||k + q||~||x_i||} \right\vert^{p}$. 

\begin{equation}
    SCS(k, x_i) = sign(k \cdot x_i) \bigg(\frac{k \cdot x_i}{(||k|| + q)(||x_i + q||)} \bigg)
    \label{eq2}
\end{equation}

Similar to the traditional convolutional process in Deep Learning, SCS is also a stride operation that takes features as an output from an input image patch. However, before calculating the straight dot product, the kernel and image patch are adjusted to have a magnitude of 1 that produce the sharpened cosine similarity or some literature refers to it as sharpened cosine normalization. SCS operation extracts feature from image patches for which sharpening exponents must be learned for each unit which can be sharpened by raising it to a power of an exponent. The aforementioned procedure outperformed and is faster than a convolutional process in any neural network by 10 to 100 times due to the fewer number of required parameters, without the normalization and activation function.

Apart from the SCS process, as compared to the pooling that is used to downsample the data, absolute max-pooling is used to update the filter in backpropagation until an appropriate value of features is obtained from the given patch. Moreover, as compared to simple pooling, absolute maxpooling chooses the highest magnitude even if a certain element has a negative value. 

\begin{figure}[!hbt]
\centering
\includegraphics[width=0.48\textwidth]{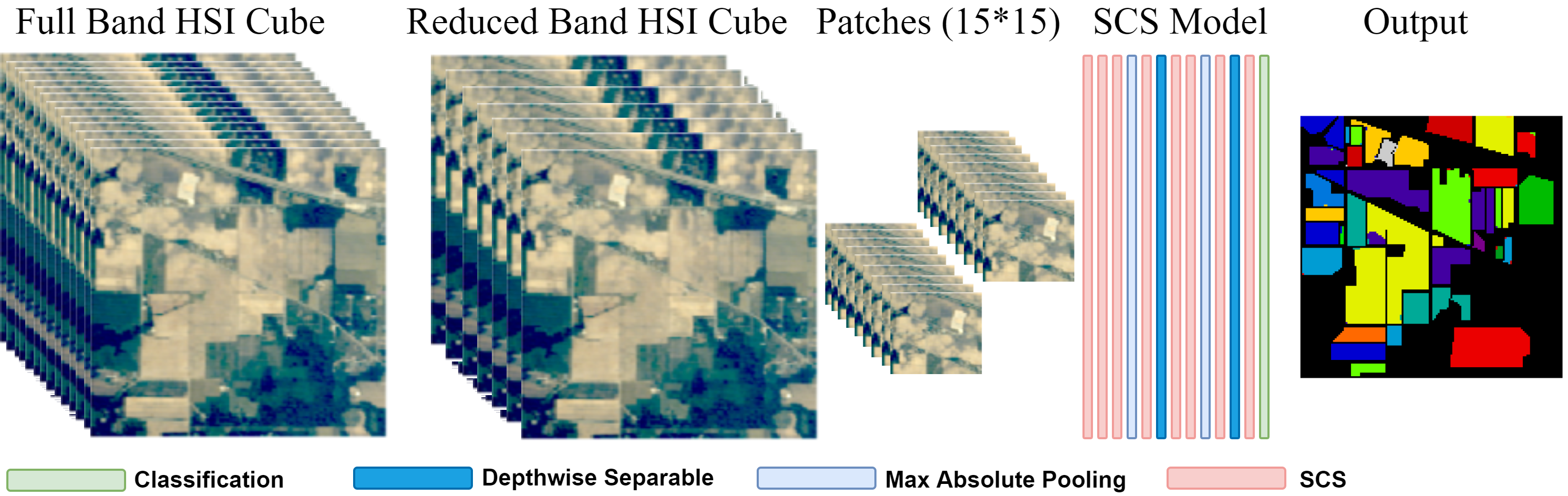}
\caption{Workflow of SCS pipeline.}
\label{fig:scatter}
\end{figure}

The overall model is trained using 250 epochs, 256 batch sizes, and a 0.001 learning rate. The learning rate has a strong influence on the pace of learning any Deep Model that determines the number of movements required to minimize the loss of function value whereas the momentum is used to enhance the accuracy and model’s training speed. The RMS and momentum prop optimizer combined model is trained using the Adam optimizer. TensorFlow-based Keras tuner library is called to fine-tune the hyperparameters for SCS-Net which ultimately makes the overall model robust and comparative as compared to the other state-of-the-art models. The motivation behind using the Adam optimizer is its efficiency in consuming less memory and computational efficiency. The complete architecture is presented in Figure \ref{fig:scatter}.

\section{Experimental Results and Discussion}
\label{Res}

This section discusses the experimental datasets used to evaluate the proposed pipeline along with the comparative methods proposed in recent years. Moreover, this section presents the details regarding the experimental settings, results, and discussion together with the experimental results as compared to the state-of-the-art methods proposed in recent years. For experimental evaluation, two publicly available HSI datasets have been used. These datasets were acquired at different times, locations, and sensors.

The works published in the literature present comprehensive results to highlight the pros and cons as compared to state-of-the-art works however, to some extent, all the works may have different experimental settings such as, the number or percentage of training, validation, and test samples may remain the same but their geographical locations may differ (due to the random selection nature) as these methods may have run in different times or may have executed on different machines. Therefore, to make the comparison fair among different works, one must need to have the same number/percentage and geographical locations of training, validation, and test samples. Thus, in this work, the first experimental settings, i.e., the percentage of training, validation, and test samples along with their geographical locations remain the same as all the comparative methods along with the SCS pipeline are executed in one single run. 

\begin{figure}[!hbt]
    \centering
    \begin{subfigure}{0.24\textwidth}
	    \centering
		\includegraphics[width=0.99\textwidth]{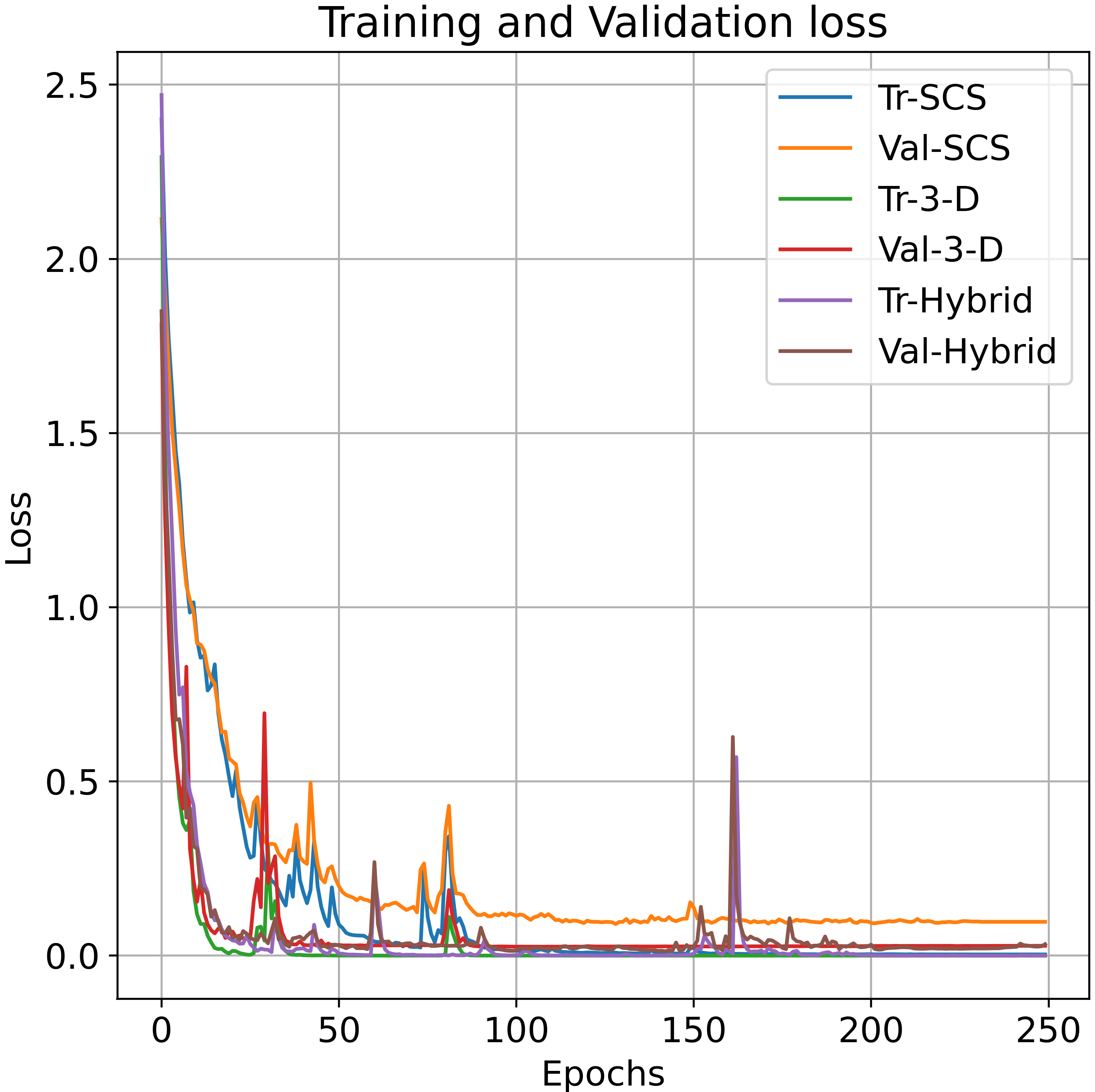}
		\caption{Loss} 
		\label{FigA1}
	\end{subfigure}
	\begin{subfigure}{0.24\textwidth}
	    \centering
		\includegraphics[width=0.99\textwidth]{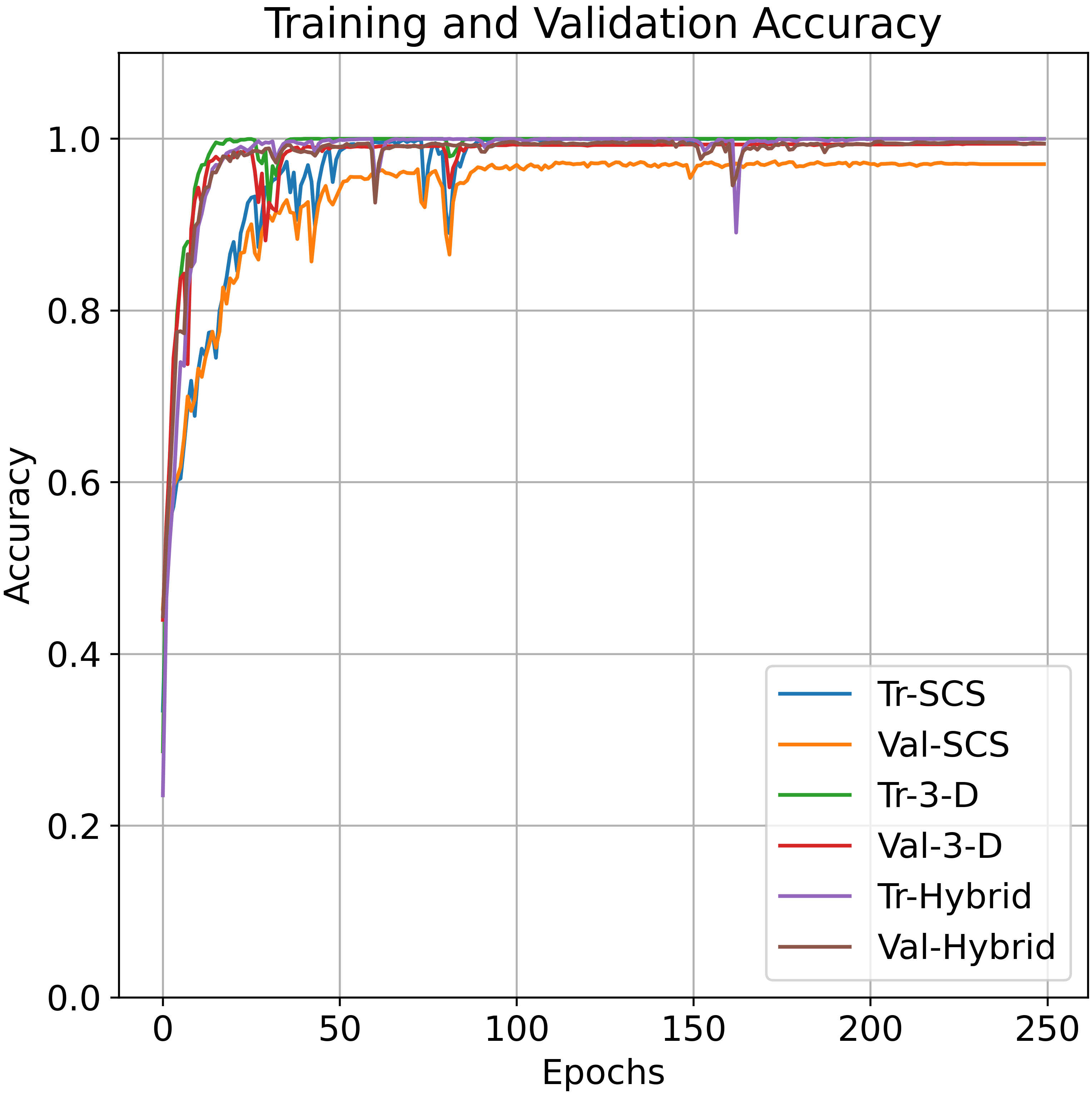}
		\caption{Accuracy}
		\label{FigA2}
	\end{subfigure}
	\caption{Training and Validation Loss and Accuracy of SCS network along with 3D-CNN and Hybrid CNN.}
	\label{FigA}
\end{figure}

The experimental results presented in this section have been obtained using Google Colab with a Graphical Process Unit with 358$+$ GB of storage and $25GB$ of RAM. For all the experimental results presented in this section the training, validation, and test samples are randomly selected as 40\%/30\%/30\%. For fair comparative analysis, all three models have been executed at once with one-time randomly selected samples. The presented results are obtained using $15 \times 15$ patch size along with the $15$ most informative bands selected through PCA. As far as the training parameters for 3D CNN and Hybrid models, the weights are initially randomized and later optimized using backpropagation with Adam optimizer and softmax loss function. The overall training and validation loss and accuracy for all three models are presented in Figure \ref{FigA}.

\begin{figure}[!hbt]
    \centering
    \begin{subfigure}{0.08\textwidth}
	    \centering
		\includegraphics[width=0.99\textwidth]{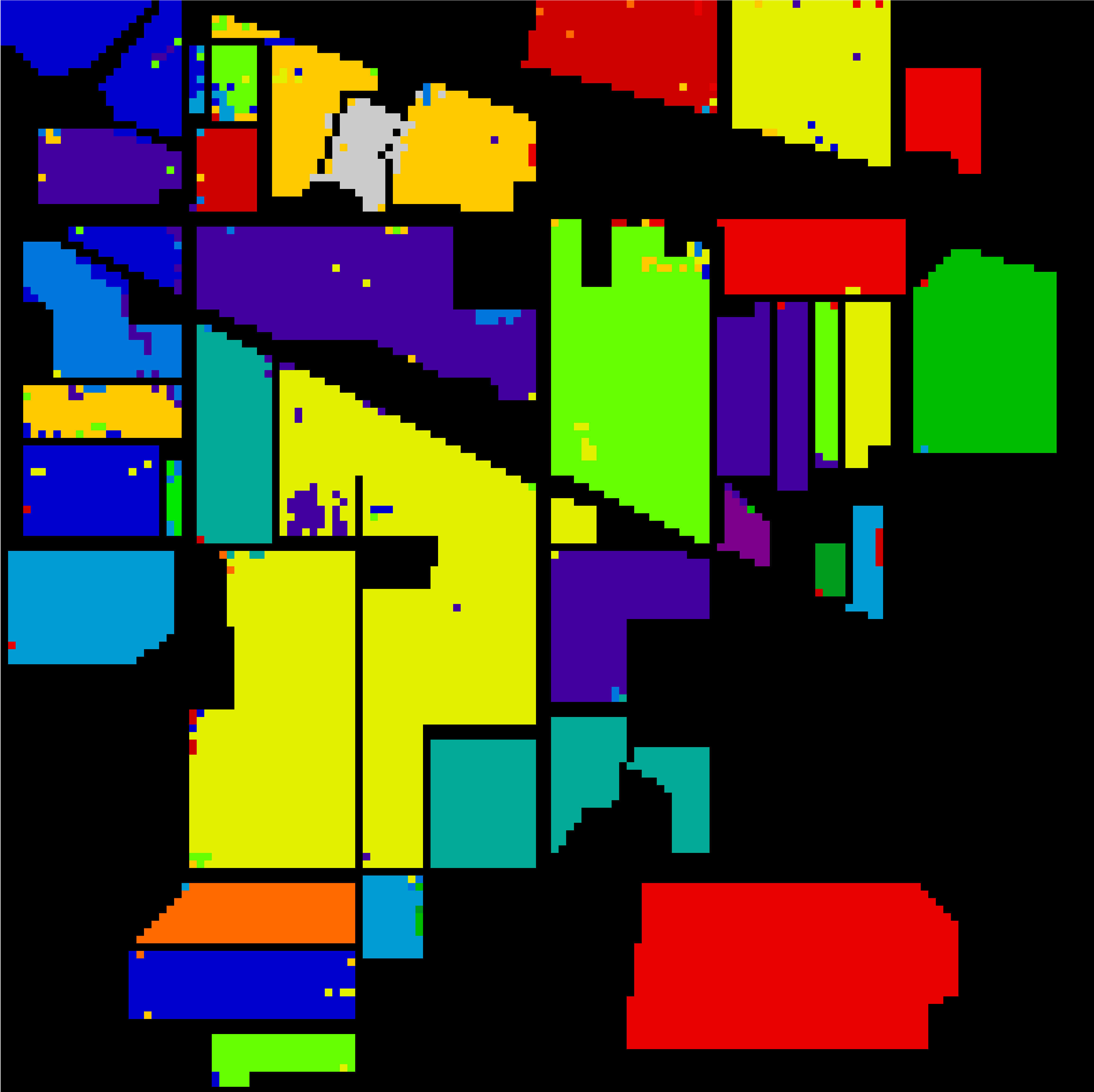}
		\caption{SCS} 
		\label{Fig2A}
	\end{subfigure}
	\begin{subfigure}{0.08\textwidth}
	    \centering
		\includegraphics[width=0.99\textwidth]{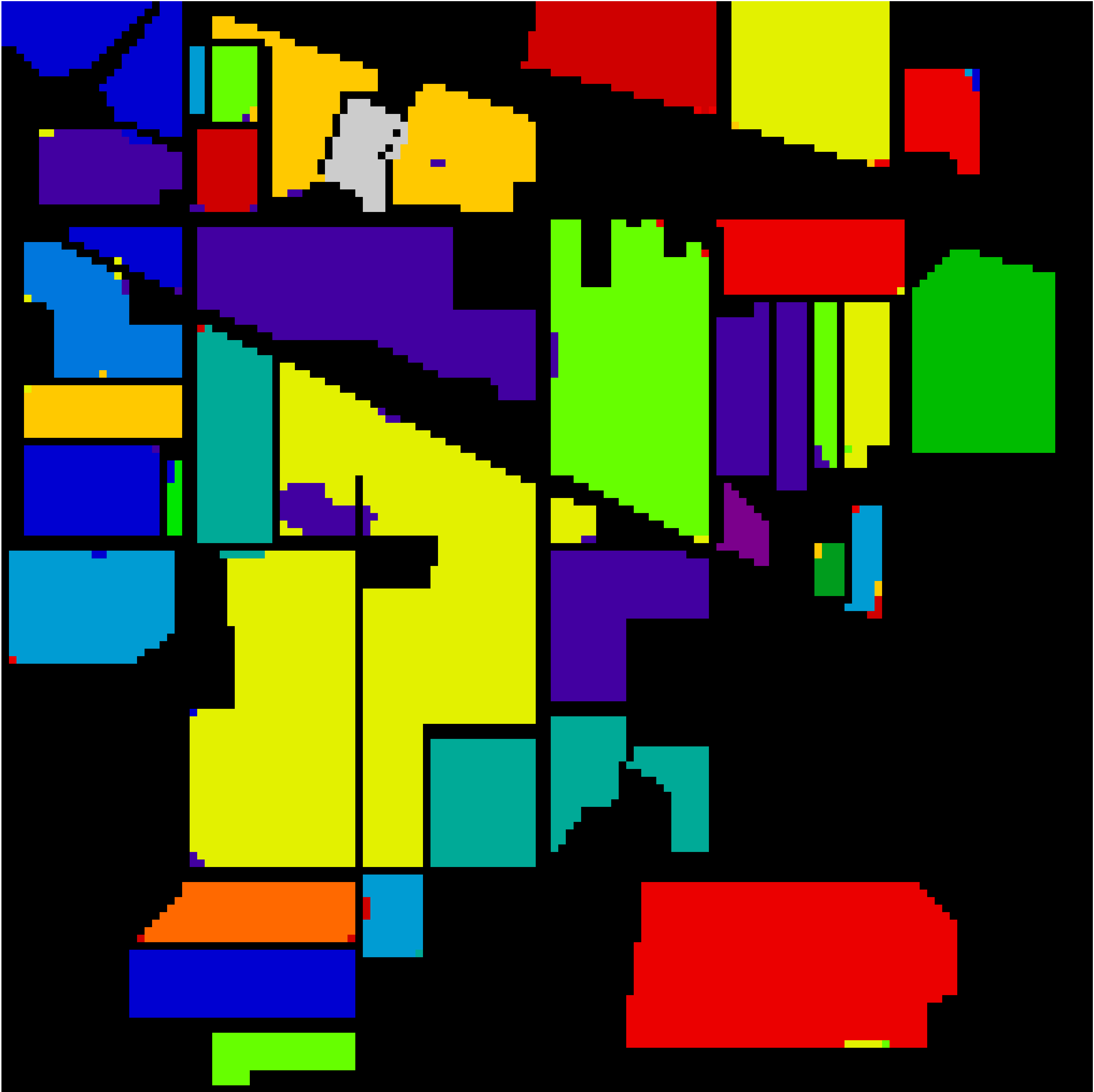}
		\caption{3D}
		\label{Fig2B}
	\end{subfigure}
	\begin{subfigure}{0.08\textwidth}
	    \centering
		\includegraphics[width=0.99\textwidth]{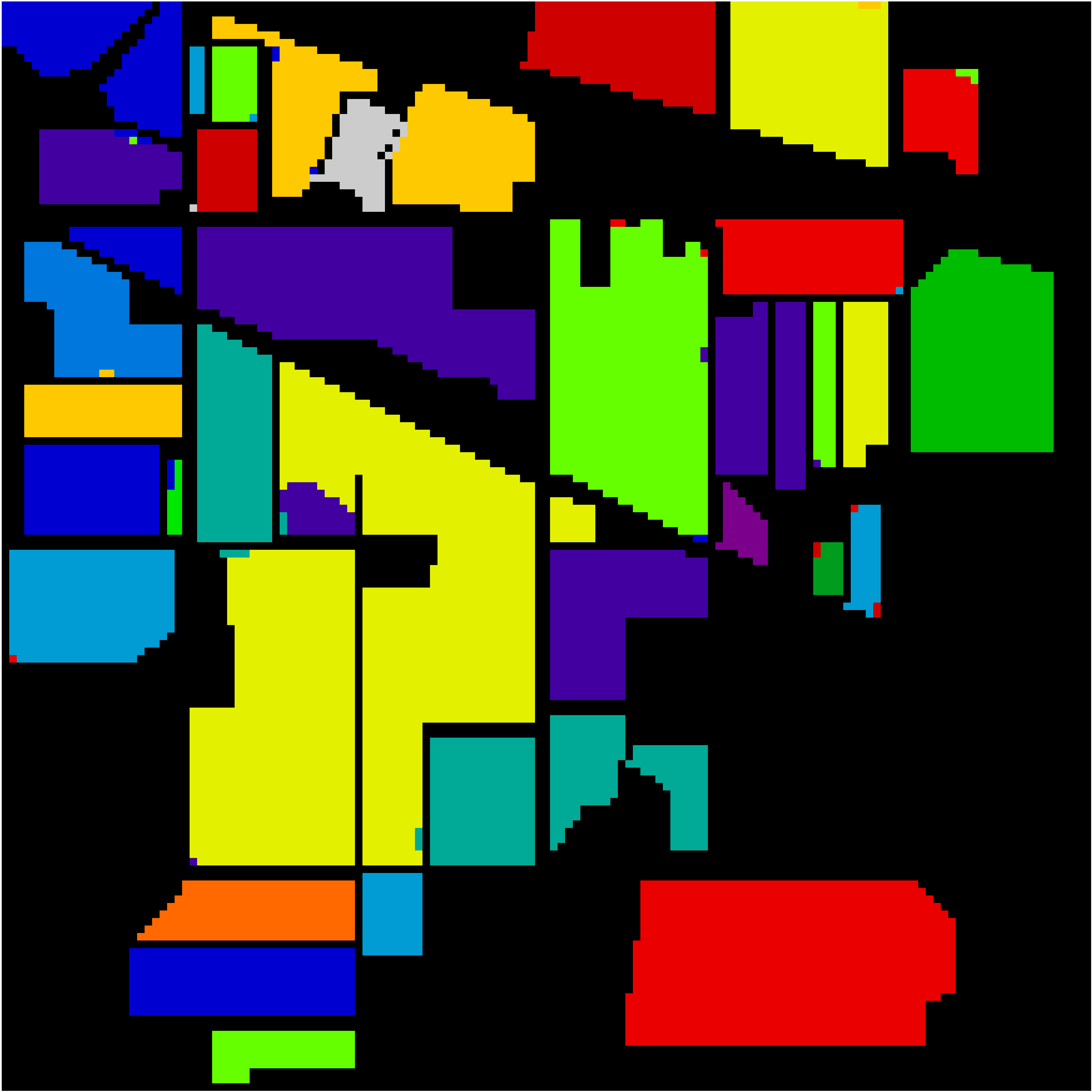}
		\caption{Hybrid} 
		\label{Fig2C}
	\end{subfigure}
    \begin{subfigure}{0.07\textwidth}
	    \centering
		\includegraphics[width=0.70\textwidth]{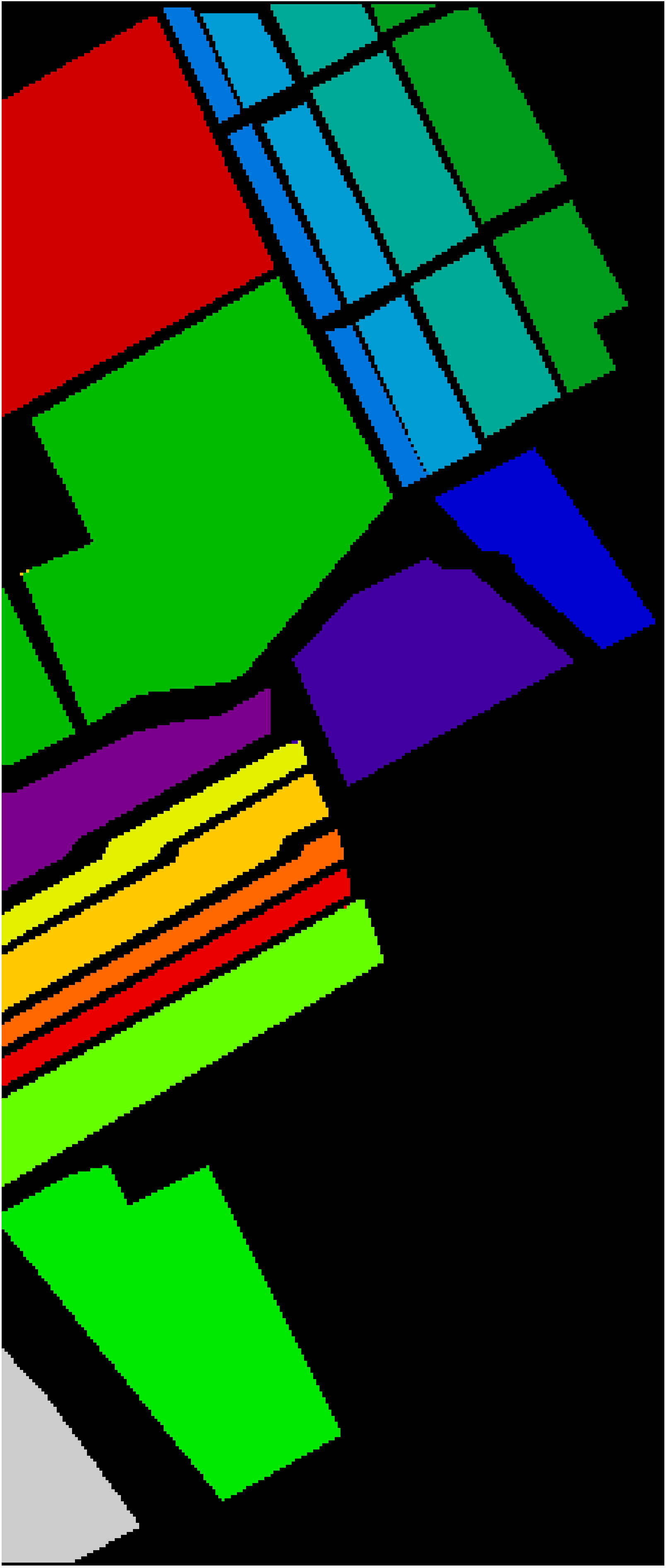}
		\caption{SCS} 
		\label{Fig3D}
	\end{subfigure}
	\begin{subfigure}{0.07\textwidth}
	    \centering
		\includegraphics[width=0.70\textwidth]{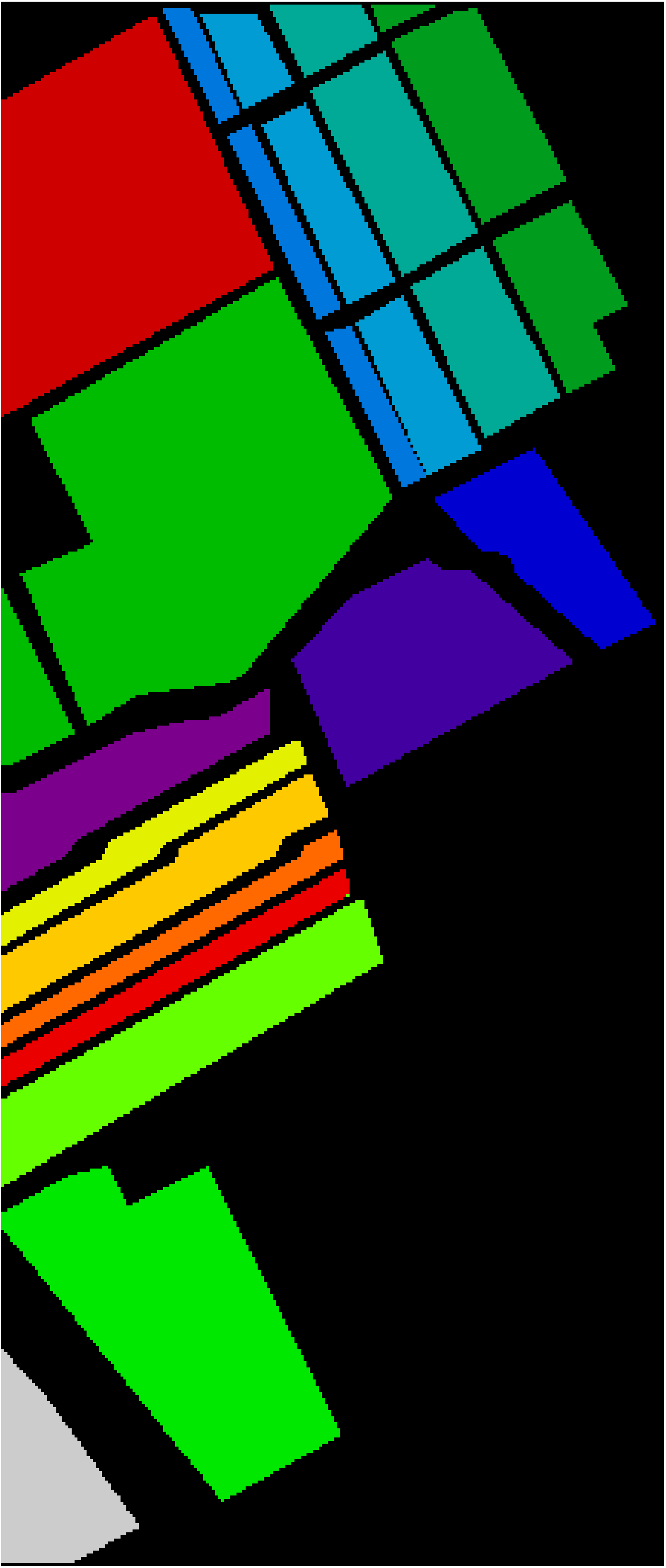}
		\caption{3D}
		\label{Fig3E}
	\end{subfigure}
	\begin{subfigure}{0.075\textwidth}
	    \centering
		\includegraphics[width=0.70\textwidth]{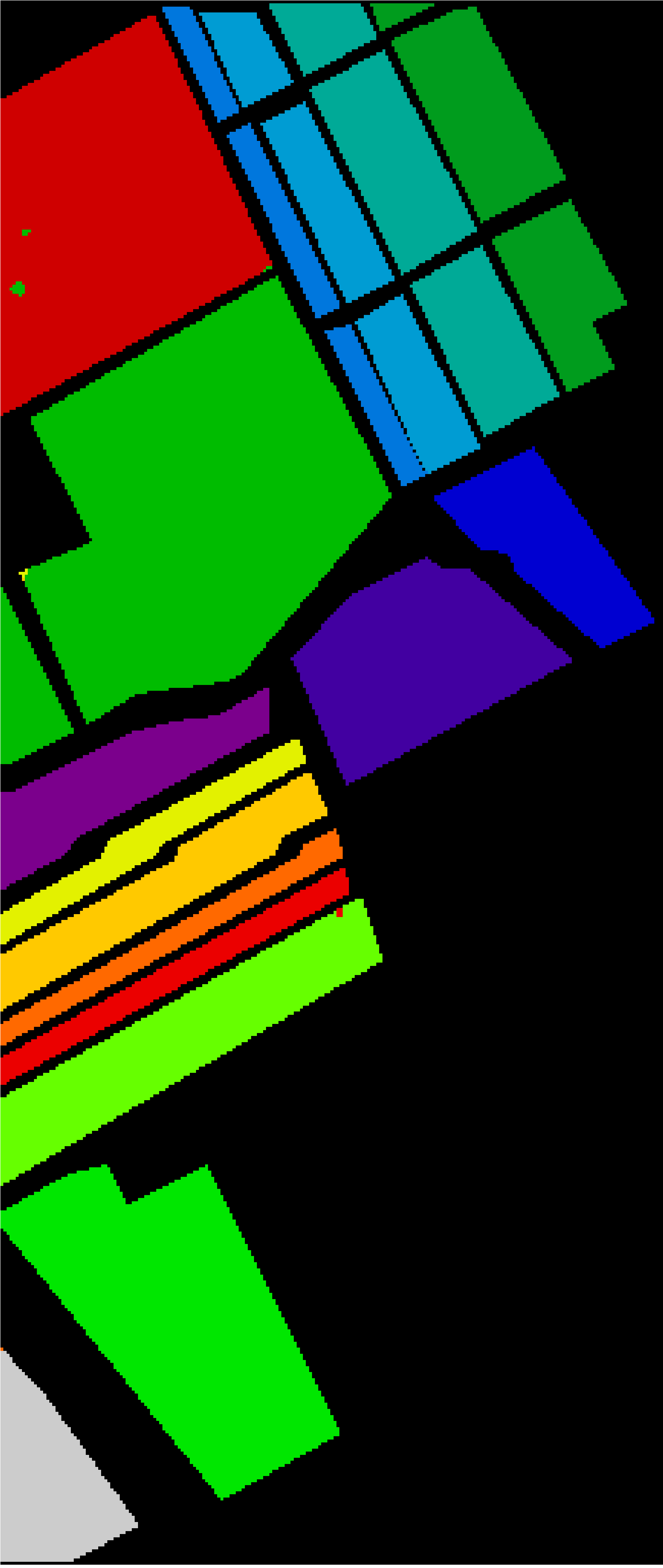}
		\caption{Hybrid} 
		\label{Fig3F}
	\end{subfigure}
	\caption{Geographical maps for Indian Pines (IP) and Salinas (SA) datasets. Per-class results are presented in Table \ref{Tab4}.}
	\label{Fig2}
\end{figure}

The experimental results presented in Figure \ref{Fig2} and Table \ref{Tab4} for both IP and SA datasets are obtained using the overall, average accuracy, and Kappa coefficient. Kappa computes the mutual information regarding a strong agreement among the classification and ground-truth maps whereas Average and Overall accuracy compute the average class-wise classification performance and the correctly classified samples out of the total test samples, respectively.

\begin{table}[!hbt] 
    \centering
    \caption{Classification performance obtained over $15 \times 15$ patch sizes.}
    \resizebox{\columnwidth}{!}{\begin{tabular}{c|c|c|c||c|c|c|c} \hline 
    \multicolumn{4}{c||}{\textbf{Indian Pines}} & \multicolumn{4}{c}{\textbf{Salinas}} \\ \hline 
    \textbf{Class} & \textbf{SCS} & \textbf{3D CNN} & \textbf{Hybrid} & \textbf{Class} & \textbf{SCS} & \textbf{3D CNN} & \textbf{Hybrid}\\ \hline 
    Alfalfa & 92.5925 & 100 & 100 & Weeds 1 & 100 & 100 & 100 \\ \hline 
    Corn-notill & 93.2322 & 97.1995 & 97.0828 & Weeds 2 & 100 & 100 & 100 \\ \hline 
    Corn-mintill & 96.5863 & 99.5983 & 100 & Fallow & 100 & 100 & 100 \\ \hline
    Corn & 84.5070 & 97.1830 & 99.2957 & Fallow RP & 100 & 100 & 100  \\ \hline 
    Grass-pasture & 93.7931 & 95.8620 & 98.9655 & Fallow smooth & 99.9253 & 100 & 99.9253 \\ \hline
    Grass-trees & 99.0867 & 99.7716 & 100 & Stubble & 100 & 100 & 100  \\ \hline
    Grass-mowed & 94.1176 & 88.2352 & 88.2352 & Celery & 100 & 100 & 100 \\ \hline 
    Hay-windrowed & 99.3031 & 100 & 100 & Grapes untrained & 99.9645 & 100 & 99.8935 \\ \hline
    Oats & 66.6666 & 75.0000 & 66.6666 & Soil vinyard develop & 100 & 100 & 100 \\ \hline
    Soybean-notill & 93.3104 & 97.5986 & 98.6277 & Corn Weeds & 99.9389 & 100 & 99.7559 \\ \hline 
    Soybean-mintill & 98.3706 & 98.7101 & 99.5926 & Lettuce 4wk & 99.6254 & 100 & 100 \\ \hline 
    Soybean-clean & 87.9213 & 97.4719 & 97.7528 & Lettuce 5wk & 100 & 100 & 100 \\ \hline  
    Wheat & 99.1869 & 98.3739 & 100 &  Lettuce 6wk & 100 & 100 & 100\\ \hline 
    Woods & 99.8682 & 98.9459 & 99.3412 & Lettuce 7wk & 100 & 99.8130 & 100 \\ \hline 
    Buildings & 95.2586 & 99.1379 & 100 & Vinyard untrained & 100 & 100 & 99.4771 \\ \hline 
    Stone-Steel & 92.8571 & 98.2142 & 100 & Vinyard trellis & 100 & 100 & 100 \\ \hline 
    \textbf{Kappa} & 95.3421 & 98.0899 & 98.8505 & \textbf{Kappa} & 99.9753 & 99.9958 & 99.8765 \\ \hline 
    \textbf{Overall} & 95.9186 & 98.3252 & 98.9918 & \textbf{Overall} & 99.9778 & 99.9963 & 99.8891 \\ \hline
    \textbf{Average} & 92.9161 & 96.3314 & 96.5975 & \textbf{Average} & 99.9658 & 99.9883 & 99.9407\\ \hline 
    \end{tabular}}
    \label{Tab4}
\end{table}

\begin{table*}[!hbt]
    \centering
    \caption{Salinas Dataset: The comparative methods include Multi-layered Perceptron (MLP) \cite{paoletti2019deep}, Multinomial Logistic Regression (MLR) \cite{li2010semisupervised}, Random Forest (RF) \cite{zhang2019active}, Support Vector Machine (SVM) \cite{melgani2004classification}, 1D CNN \cite{hong2021graph}, 2D CNN \cite{makantasis2015deep}, 3D CNN \cite{ahmad2020fast}, Hybrid CNN \cite{ahmad2021hyperspectral}, Bayesian CNN (1D, 2D, and 3D BCNN) \cite{8390931}. All these methods are retrained using the same number of training samples to make the comparison fair and reliable.}
    
    \resizebox{\textwidth}{!}{\begin{tabular}{c|c|c|c|c|c|c|c|c|c|c|c||c} \hline
        Class & MLP & MLR & RF & SVM & 1D CNN & 2D CNN & 3D CNN & Hybrid CNN & 1D-BCNN & 2D-BCNN & 3D-BCNN & SCS \\ \hline
        
        Broccoli green weeds 1 & 98.16$\pm$1.94 & 98.26$\pm$0.61 & 97.71$\pm$1.94 & 97.59$\pm$1.31 & 99.00$\pm$0.42 & 99.85$\pm$0.15 & 100$\pm$0.00 & 100$\pm$0.00 & 99.8 $\pm$0.22 & 99.55 $\pm$0.45 & 100 $\pm$0.00 & 100$\pm$0.00 \\ \hline 
        
        Broccoli green weeds 2 & 99.48$\pm$0.40 & 99.78$\pm$0.07 & 99.83$\pm$0.07 & 99.35$\pm$0.45 & 99.55$\pm$0.00 & 94.15$\pm$1.45 & 100$\pm$0.00 & 100$\pm$0.00 & 99.97 $\pm$0.02 & 99.72 $\pm$0.28 & 100 $\pm$0.00 & 100$\pm$0.00 \\ \hline 
        
        Fallow & 96.89$\pm$1.76 & 94.94$\pm$1.82 & 93.74$\pm$3.59 & 96.88$\pm$2.08 & 97.79$\pm$0.56 & 99.62$\pm$0.03 & 100$\pm$0.00 & 100$\pm$0.00 & 99.68 $\pm$0.21 & 99.89 $\pm$0.11 & 100 $\pm$0.00 & 100$\pm$0.00 \\ \hline 
        
        Fallow rough plow & 99.44$\pm$0.31 & 99.24$\pm$0.38 & 97.06$\pm$3.00 & 98.98$\pm$0.61 & 98.76$\pm$0.96 & 99.86$\pm$0.14 & 100$\pm$0.00 & 100$\pm$0.00 & 99.71 $\pm$0.01 & 99.89 $\pm$0.11 & 99.61 $\pm$0.18 & 100$\pm$0.00 \\ \hline 
        
        Fallow smooth & 97.50$\pm$1.15 & 97.36$\pm$1.21 & 96.25$\pm$0.99 & 97.87$\pm$0.72 & 96.98$\pm$1.18 & 99.79$\pm$0.06 & 100$\pm$0.00 & 99.93$\pm$0.07 & 98.51 $\pm$1.51 & 100 $\pm$0.00 & 99.91 $\pm$0.06 & 99.92$\pm$0.08 \\ \hline
        
        Stubble & 99.52$\pm$0.22 & 99.57$\pm$0.18 & 98.73$\pm$0.99 & 99.43$\pm$0.40 & 99.80$\pm$0.13 & 99.73$\pm$0.21 & 100$\pm$0.00 & 100$\pm$0.00 & 99.97 $\pm$0.00 & 100 $\pm$0.00 & 100 $\pm$0.00 & 100$\pm$0.00 \\ \hline 
        
        Celery & 99.27$\pm$0.33 & 99.66$\pm$0.16 & 99.09$\pm$0.41 & 99.44$\pm$0.21 & 99.68$\pm$0.09 & 99.09$\pm$0.15 & 100$\pm$0.00 & 100$\pm$0.00 & 99.97 $\pm$0.00 & 100 $\pm$0.00 & 100 $\pm$0.00 & 100$\pm$0.00 \\ \hline 
        
        Grapes untrained & 81.16$\pm$5.33 & 81.89$\pm$3.01 & 81.85$\pm$2.60 & 97.53$\pm$1.78 & 83.43$\pm$3.15 & 92.31$\pm$1.01 & 100$\pm$0.00 & 99.89$\pm$0.00 & 86.61 $\pm$5.95 & 9997 $\pm$0.03 & 99.88 $\pm$0.12 & 99.96$\pm$0.04 \\ \hline 
        
        Soil vinyard develop & 99.34$\pm$0.43 & 99.86$\pm$0.07 & 98.90$\pm$0.44 & 99.39$\pm$0.52 & 99.26$\pm$0.43 & 99.84$\pm$0.06 & 100$\pm$0.00 & 100$\pm$0.00 & 99.96 $\pm$0.01 & 100 $\pm$0.00 & 100 $\pm$0.00 & 100$\pm$0.00 \\ \hline 
        
        Corn senesced green weeds & 89.33$\pm$2.19 & 88.50$\pm$2.12 & 85.53$\pm$1.96 & 91.13$\pm$1.74 & 93.49$\pm$2.15 & 96.19$\pm$2.81 & 100$\pm$0.00 & 99.76$\pm$0.24 & 98.59 $\pm$0.48 & 99.82 $\pm$0.18 & 100 $\pm$0.00 &99.93$\pm$0.07 \\ \hline 
        
        Lettuce romaine 4wk & 90.02$\pm$3.76 & 91.95$\pm$3.05 & 88.16$\pm$4.53 & 93.93$\pm$1.83 & 94.48$\pm$1.99 & 96.82$\pm$0.84 &  100$\pm$0.00 & 100$\pm$0.00 & 98.94 $\pm$0.94 & 100 $\pm$0.00 & 100 $\pm$0.00 & 99.62$\pm$0.38 \\ \hline 
        
        Lettuce romaine 5wk & 97.21$\pm$2.40 & 99.03$\pm$0.73 & 97.19$\pm$1.37 & 99.14$\pm$0.56 & 99.97$\pm$0.05 & 99.82$\pm$0.18 & 100$\pm$0.00 & 100$\pm$0.00 & 99.48 $\pm$0.24 & 99.92 $\pm$0.08 & 99.91 $\pm$0.03 & 100$\pm$0.00 \\ \hline 
        
        Lettuce romaine 6wk & 97.66$\pm$1.32 & 94.39$\pm$8.09 & 97.79$\pm$1.37 & 97.39$\pm$2.38 & 98.25$\pm$0.62 & 98.42$\pm$1.15 & 100$\pm$0.00 & 100$\pm$0.00 & 99.49 $\pm$0.19 & 100 $\pm$0.00 & 99.95 $\pm$0.05 & 100$\pm$0.00 \\ \hline 
        
        Lettuce romaine 7wk & 91.38$\pm$2.33 & 92.26$\pm$1.34 & 90.88$\pm$3.21 & 91.92$\pm$3.07 & 91.03$\pm$1.75 & 96.82$\pm$0.00 & 100$\pm$0.00 & 100$\pm$0.00 & 99.19 $\pm$0.23 & 99.63 $\pm$0.09 & 100 $\pm$0.00 & 100$\pm$0.00 \\ \hline 
        
        Vinyard untrained & 64.87$\pm$8.76 & 60.89$\pm$3.55 & 59.21$\pm$4.36 & 64.20$\pm$2.91 & 66.41$\pm$7.54 & 84.74$\pm$1.35 & 99.81$\pm$0.19 & 99.48$\pm$0.52 & 74.95 $\pm$9.27 & 99.64 $\pm$0.30 & 99.66 $\pm$0.34 & 100$\pm$0.00 \\ \hline 
        
        Vinyard vertical trellis & 96.36$\pm$1.20 & 95.29$\pm$2.48 & 92.92$\pm$2.26 & 96.70$\pm$1.84 & 98.34$\pm$0.57 & 85.78$\pm$0.39 & 100$\pm$0.00 & 100$\pm$0.00 & 99.56 $\pm$0.24 & 99.81 $\pm$0.19 & 100 $\pm$0.00 & 100$\pm$0.00 \\ \hline \hline
        
        \textbf{OA} & 89.57$\pm$0.41 & 89.20$\pm$0.30 & 88.22$\pm$0.29 & 91.07$\pm$0.37 & 90.85$\pm$0.77 & 94.95$\pm$0.07 & 99.99$\pm$0.01 & 99.88$\pm$0.18 & 93.57 $\pm$0.29 & 99.88 $\pm$0.08 & 99.91 $\pm$0.08 & 99.97$\pm$0.03 \\ \hline
        
        \textbf{AA} & 93.60$\pm$0.56 & 93.30$\pm$0.59 & 92.18$\pm$0.28 & 94.43$\pm$0.38 & 94.79$\pm$0.64 & 96.43$\pm$0.23 & 99.98$\pm$0.02 & 99.94$\pm$0.06 & 97.57 $\pm$0.33 & 99.87 $\pm$0.07 & 99.94 $\pm$0.05 & 99.96$\pm$0.04 \\ \hline
        
        \textbf{$\kappa$} & 88.38$\pm$0.47 & 89.20$\pm$0.33 & 86.86$\pm$0.33 & 90.03$\pm$0.41 & 90.85$\pm$0.87 & 94.95$\pm$0.08 & 99.99$\pm$0.01 & 99.87$\pm$0.13 & 93.57 $\pm$0.32 & 99.88 $\pm$0.09 & 99.91 $\pm$0.09 & 99.97$\pm$0.03\\ \hline
    \end{tabular}}
    \label{Tab10}
\end{table*}

\section{Experiments with State-of-the-art Models}

The section discussed the experimental results of SCS and comparative methods proposed in recent years. This work carryout the comparative results with the same settings, irrespective of the literature, e.g., training and test samples may have the same number or percentage but are selected in different runs which leads to the different geographical locations of training and test samples for different methods. For instance, some samples are close to the decision boundary and are hard to classify (may confuse classification) however, in another random selection run, such samples may not be a part of the training set eventually, making classification easy. Therefore, all the comparative methods must need to be evaluated on the same set of samples rather than different samples. In this work, the training and test samples are selected once only and used to train and test all the models for the same samples. 

Here the SCS method is compared with Multi-layered Perceptron (MLP) \cite{paoletti2019deep}, Multinomial Logistic Regression (MLR) \cite{li2010semisupervised}, Random Forest (RF) \cite{zhang2019active}, Support Vector Machine (SVM) \cite{melgani2004classification}, 1D CNN \cite{hong2021graph}, 2D CNN \cite{makantasis2015deep}, 3D CNN \cite{ahmad2020fast}, Hybrid CNN \cite{ahmad2021hyperspectral}, Bayesian CNN (1D, 2D, and 3D BCNN) \cite{8390931}. All these methods were implemented as per the parameters mentioned in their respective works. The detailed experimental results are presented in Table \ref{Tab10} for the Salains dataset (due to the page length, other datasets' results are skipped). Looking into Table \ref{Tab10}, one can conclude that the pixel-based classification methods i.e., MLR and RF under-performed SVM, however, 1D CNNs performance is superior to other pixel-based spectral classification methods. Whereas, the spatial and spectral-spatial classification methods such as 2D, 3D, and Hybrid methods, respectively outperformed spectral methods. Moreover, SCS performance is similar to spectral-spatial methods' performance while requiring significantly fewer training parameters, i.e., SCS requires $5624$ training parameters whereas, 3D and Hybrid CNN require $127,104$ training parameters, which is significant.

\section{Conclusion}
\label{Con}

Hyperspectral Image Classification is a difficult task due to high inter and intra-class similarity and variability, nested regions, and overlapping. 2D Convolutional Neural Networks (CNN) emerged as a viable network whereas, 3D CNNs are a better alternative due to accurate classification. However, 3D CNNs are highly computationally complex due to their volume and spectral dimensions. Moreover, down-sampling and hierarchical filtering (high frequency) i.e., texture features need to be smoothed during the forward pass which is crucial for accurate classification. Furthermore, CNN requires tons of tuning parameters which increases the training time. To overcome the aforesaid issues, this work presented a Sharpened Cosine Similarity (SCS) as an alternative to Convolution in neural networks. SCS is exceptionally parameter efficient due to the following reasons: Skip nonlinear activation layer (ReLU, Sigmoid, etc.), Skip normalization and dropout layers, Use of Abs Max-pooling instead of MaxPool which selects the elements with the highest magnitude of activity even if its negative. Several experimental results proved the efficiency of SCS as compared to Convolutional layers.

\bibliographystyle{IEEEtran}
\bibliography{IEEEabrv,Sam}
\end{document}